%% file: main.tex
\pdfoutput=1
\documentclass{article}

\usepackage[preprint]{neurips_2026}
\makeatletter
\renewcommand{\@noticestring}{40th Conference on Neural Information Processing Systems (NeurIPS 2026). Workshop: 2nd Workshop on Embodied Spatial Reasoning (ESR).}
\makeatother

\usepackage[utf8]{inputenc}
\usepackage[T1]{fontenc}
\usepackage{url}
\usepackage{booktabs}
\usepackage{amsfonts}
\usepackage{amsmath}
\usepackage{amssymb}
\usepackage{nicefrac}
\usepackage{microtype}
\usepackage{xcolor}
\usepackage{graphicx}
\usepackage{float}
\usepackage{tikz}
\usepackage{pifont}
\usetikzlibrary{arrows.meta, positioning, calc}
\usepackage[hidelinks]{hyperref}

\newcommand{\verify}[1]{#1}

\title{ SpaceMCP: Vision Language Model Spatial Reasoning via Model Context Protocol }

\author{%
  Ameen Izhac \quad Nobuko Yoshida \quad Ronald Clark \\
  Department of Computer Science \\
  University of Oxford \\
}

\begin{document}

\maketitle

\begin{figure}[H]
\centering
\resizebox{0.92\linewidth}{!}{\input{fig1_reasoning.tex}}
\caption{\textbf{Inside the VLM's head: SpaceMCP swaps guessing for
grounding.} An illustrative ego-perspective question over an indoor
video. Direct VLM \emph{(right)} narrates uncertainty and guesses.
SpaceMCP \emph{(middle)} routes a short tool-call sequence through
MCP, lifts a metric answer (62\textdegree) out of the scene graph,
and converts it into the cardinal answer (call names abbreviated).}
\label{fig:reasoning}
\end{figure}
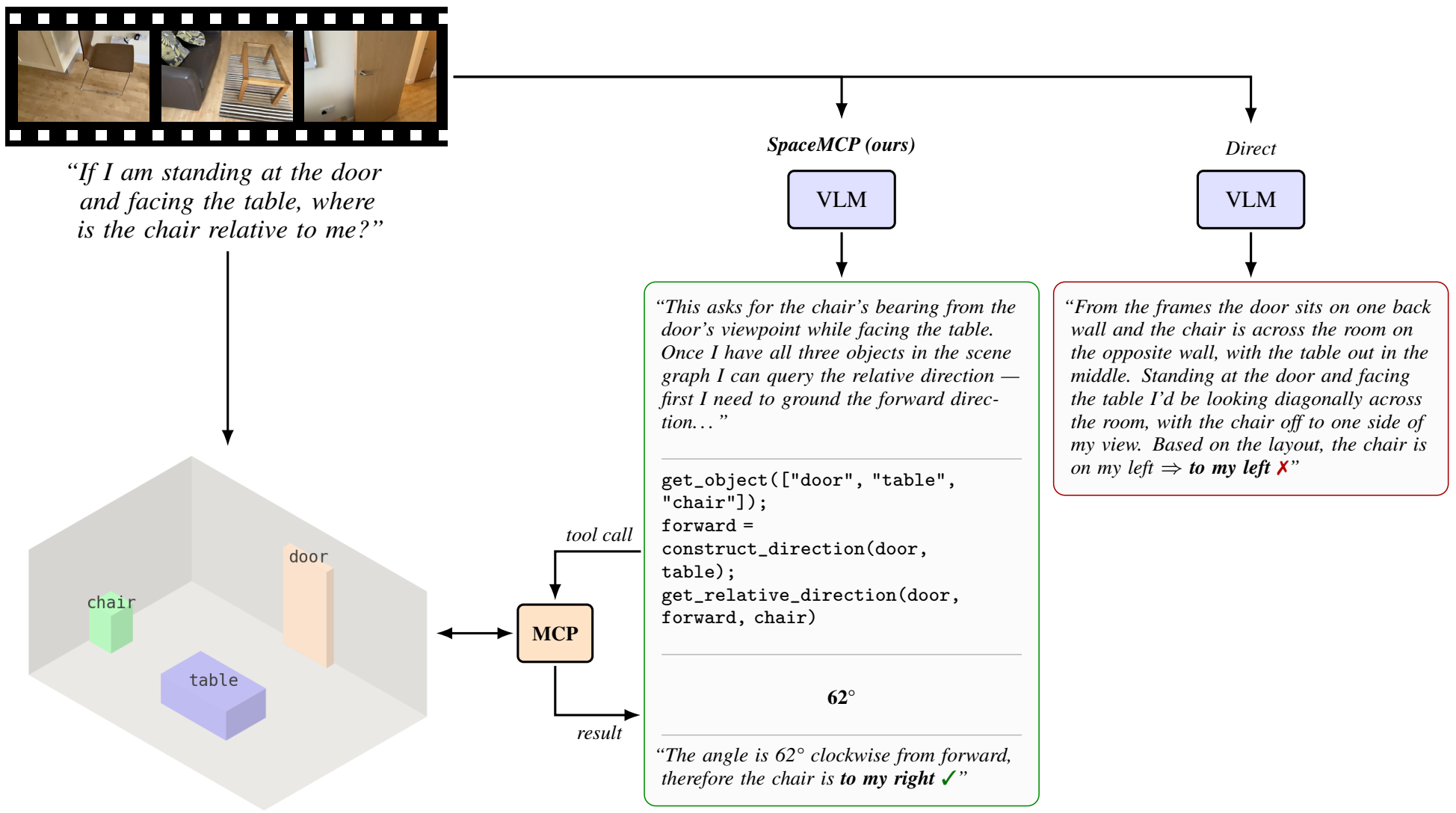

\begin{abstract}
Despite impressive advances in large language models (LLM), vision–language models (VLM) still
exhibit limited spatial reasoning capabilities. Recent progress in feed-forward 3D
reconstruction has enabled precise, real-time geometric understanding from visual
inputs. In this work, we propose SpaceMCP, a framework that leverages these advances to
introduce an MCP-style paradigm for spatial reasoning. Our approach augments
a VLM with structured function calls, allowing it to query a scene
graph for spatial information such as distances, object positioning, and camera
properties. The scene graph is constructed in real time using feed-forward 3D reconstruction
to recover metric geometry, combined with open-vocabulary segmentation to identify and localize
relevant objects within the scene. Paired with the VLM's strong reasoning capabilities, the VLM
is able to combine this primitive information to solve more complex spatial tasks.
Without any training, SpaceMCP achieves state-of-the-art results on VSI-Bench, MMSI-Bench and
VSTI-Bench, outperforming recent spatially post-trained VLMs and improving its base VLM by
15--22 points per benchmark. We argue that, given the contrast between
the relatively slow progress
in native VLM spatial reasoning and the rapid advances in feed-forward 3D computer vision,
this paradigm offers a promising path forward for enabling robust spatial reasoning.
\end{abstract}

\section{Introduction}
\label{sec:intro}

Spatial reasoning is a cornerstone of general intelligence and a
prerequisite for deploying autonomous agents in the physical world, from
robot navigation and planning \citep{navila,conceptgraphs} to AR/VR. Yet a stark
asymmetry has emerged in modern foundation models: large language
models now match or exceed human performance on broad knowledge and
reasoning benchmarks --- frontier models now score above $90\%$ on GPQA
Diamond \citep{gpqaleaderboard}, against $65\%$ for PhD-domain experts
\citep{gpqa} --- while
vision-language models (VLMs) lag dramatically when the same competence
is required in 3D space. On MMSI-Bench \citep{mmsi}, GPT-5 reaches only
$41.9\%$, and SenseNova-SI \citep{sensenova}, a frontier VLM
trained at scale specifically for spatial understanding, attains
$43.3\%$, against $97.2\%$ for humans; comparable gaps persist on
VSI-Bench \citep{vsibench}, ViewSpatial-Bench \citep{viewspatial}, and
SPACE \citep{space}, and even SpatialVLM \citep{spatialvlm},
fine-tuned on roughly two billion synthetic spatial-QA examples, trails human
performance by a wide margin. The shortfall is widely attributed to a
training distribution dominated by 2D web imagery with little complex
spatial supervision \citep{sqa3d,3dsrbench}.

Existing efforts to close this gap fall into three families
\citep{spatialsurvey25}. \emph{Post-training} on spatial-specific data
\citep{spatialvlm,spatialrgpt,mmspatial,spatialthinker,spatialladder,sensenova,euclidsgift}
teaches a VLM to internalise spatial facts from large curated corpora.
\emph{Architectural integration} of a 3D expert
\citep{vlm3r,spacemind,spatialmllm,spatialllm,inst3dlmm,s2mllm,arpvlm,threedllm}
fuses geometric features from a 3D backbone into the VLM at training
time. \emph{Training-free methods}
\citep{mindjourney,visualagentic,tiger,mssr,spatialdreamer,spatialbot,eaglevision,seeandtrek}
leave the VLM's weights untouched and instead route inputs through
external tools (tool augmentation) or expanded inference loops
(test-time scaling).

Training-free tool augmentation is the closest in spirit to our work,
but existing instances are tied to a single question type: SpatialBot
\citep{spatialbot} exposes a depth-query API that solves
point-distance questions but says nothing about counts, layouts or
ego-perspective bearings; SpatialDreamer \citep{spatialdreamer}
and MindJourney \citep{mindjourney} synthesise alternative viewpoints
with a novel-view or world model and help almost exclusively on
view-change questions; TIGeR \citep{tiger} wires classical geometric
tools into a VLM agent; MSSR \citep{mssr} prunes a 3D toolbox to a
per-question minimal set. Each helps on its target distribution and
does nothing for adjacent question types, and several run generative
synthesis in their inner loop at inference costs orders of magnitude
above direct prompting.

Two recent advances make a simpler design viable. First, feed-forward
3D reconstructors such as VGGT \citep{vggt} and DUSt3R \citep{dust3r}
produce dense, geometrically consistent point clouds from monocular
video in a single forward pass, and recent variants --- notably Depth
Anything 3 \citep{da3} and MapAnything \citep{mapanything} --- emit
depth in metric units directly. Second, open-vocabulary segmenters such
as SAM~3 \citep{sam3} promote those reconstructions to typed
scene-graph nodes from free-form language prompts, without a closed
object vocabulary.

Building on these advances, we present \textbf{SpaceMCP}, a
training-free, tool-augmented framework that exposes a metric scene
graph to the VLM through a generic Model-Context-Protocol (MCP)
interface \citep{mcp}. From a single-room indoor video (the setting of
the benchmarks we target; Sec.~\ref{sec:limitations} discusses which
components carry that assumption), SpaceMCP reconstructs the scene
with Depth Anything 3 \citep{da3}, which supplies metric depth and
camera extrinsics in one forward pass, segments objects with SAM~3 on
demand, and derives the room layout from a 2~cm occupancy grid. The
scene graph is populated \emph{statically} on receipt of the video
(room footprint, floor plane, metric point cloud) and
\emph{dynamically}, with object instances added at the VLM's request
via two augmentation tools (\texttt{add\_object},
\texttt{add\_objects}); read-only getters expose object
dimensions, distances, relative direction, room size, first appearance, and
camera poses for ego-perspective and camera-motion questions. At
inference the VLM emits a single MCP script combining augmentations
and getter calls and answers from the response, in real time with
negligible overhead over a vanilla forward pass.

The crucial design choice is what we expose. Rather than a tool
engineered around one question type --- a novel-view renderer, a depth
probe, a world-model rollout --- we expose a small set of
\emph{primitive} metric queries (positions, sizes, distances, room
footprint, bearings, camera poses) and let the VLM compose them.
Counting, size, distance, room dimensions, ``standing-at-X facing-Y''
bearings and camera motion all reduce to short call sequences over the
same fixed surface, shifting the bottleneck of tool augmentation from
\emph{engineering a new tool per question type} to \emph{designing one
primitive interface}: new primitives can be added once and benefit
every question that decomposes into them.

Our contributions are:
\begin{itemize}
  \item \textbf{A general MCP interface for spatial reasoning.} Unlike
    prior tool-augmented work that engineers a separate tool per
    question type, we expose a small set of metric primitives over a
    real-time scene graph, and a single fixed surface covers counting,
    size, distance, layout, ego-perspective and camera-motion
    questions without per-task adaptation.
  \item \textbf{A static and dynamic scene-graph generation approach.}
    Video-level information is distilled statically on receipt (room
    size, floor plane, metric point cloud) and expanded dynamically
    via VLM-driven augmentation calls (object segmentation).
  \item \textbf{State-of-the-art across benchmarks.} Without any
    training, SpaceMCP lifts its base VLM from 55.8 to 71.8 on
    VSI-Bench \citep{vsibench}, 34.3 to 49.2 on MMSI-Bench
    \citep{mmsi}, and 49.0 to 71.1 on VSTI-Bench \citep{vlm3r},
    surpassing recent spatially post-trained VLMs on all three; a
    fully open-source 27B-parameter backbone obtains comparable gains.
\end{itemize}

\section{Related Work}
\label{sec:related}

\paragraph{Benchmarks and trained approaches.}
VSI-Bench \citep{vsibench}, ViewSpatial-Bench \citep{viewspatial},
SPACE \citep{space}, 3DSRBench \citep{3dsrbench}, MMSI-Bench
\citep{mmsi} and SpaceVista \citep{spacevista} find VLMs weak at
metric estimation, viewpoint reasoning and layout understanding alike.
Post-training methods curate synthetic spatial QA
\citep{spatialvlm,spatialrgpt,mmspatial}, add reinforcement learning
with spatial rewards \citep{spatialthinker,spacer,spatialssrl,spatialladder},
or scale the supervised recipe to 8M examples \citep{sensenova};
architectural methods fuse a 3D expert into the VLM
\citep{vlm3r,spacemind,spatialmllm,cambrian,spatialllm}.
Both inherit the cost of training and tend to overfit benchmark
formats faster than they generalise across them.

\paragraph{Training-free, tool-augmented and programmatic methods.}
AVIC \citep{avic} gates novel-view synthesis on its expected value;
ViLaSR \citep{vilasr} interleaves reasoning with on-image drawing;
TIGeR \citep{tiger} and MSSR \citep{mssr} give a VLM agent classical
geometric tools. SpatialPIN \citep{spatialpin} and Struct2D
\citep{struct2d} instead pass 3D facts as plain text or a rendered
BEV; our flattened-text ablation (Sec.~\ref{sec:attribution})
instantiates this design with identical perception outputs and shows
it recovers only part of the benefit of a callable interface. VisProg
\citep{visprog}, ViperGPT \citep{vipergpt} and VADAR
\citep{visualagentic} have an LLM emit programs over vision
sub-modules, but operate on 2D imagery or isolated frames, are
stateless across queries, and predate the feed-forward reconstruction
\citep{mast3r,vggt,da3} and open-vocabulary segmentation \citep{sam3}
that make a stateful metric scene graph viable.

\section{Method}
\label{sec:method}

\begin{figure}[t]
\centering
\resizebox{\linewidth}{!}{\input{fig2_overview.tex}}
\caption{\textbf{SpaceMCP overview.} \emph{Top:} DA3 lifts the 1\,fps
frames to a metric point cloud with per-frame camera poses; SAM~3
grounds the object names in the plan on the cached features, the
masked points are filtered and fitted with OBBs, and the boxes and the
camera path form the scene graph. \emph{Bottom:} for question $q$ the
VLM emits one plan of \texttt{add\_object} calls and getters; the
getters run against the scene graph, here
\texttt{construct\_direction} followed by
\texttt{get\_relative\_direction} on its result, and the VLM turns the
returned bearing into the answer in a second turn. Clouds and boxes
are schematic.}
\label{fig:overview}
\end{figure}
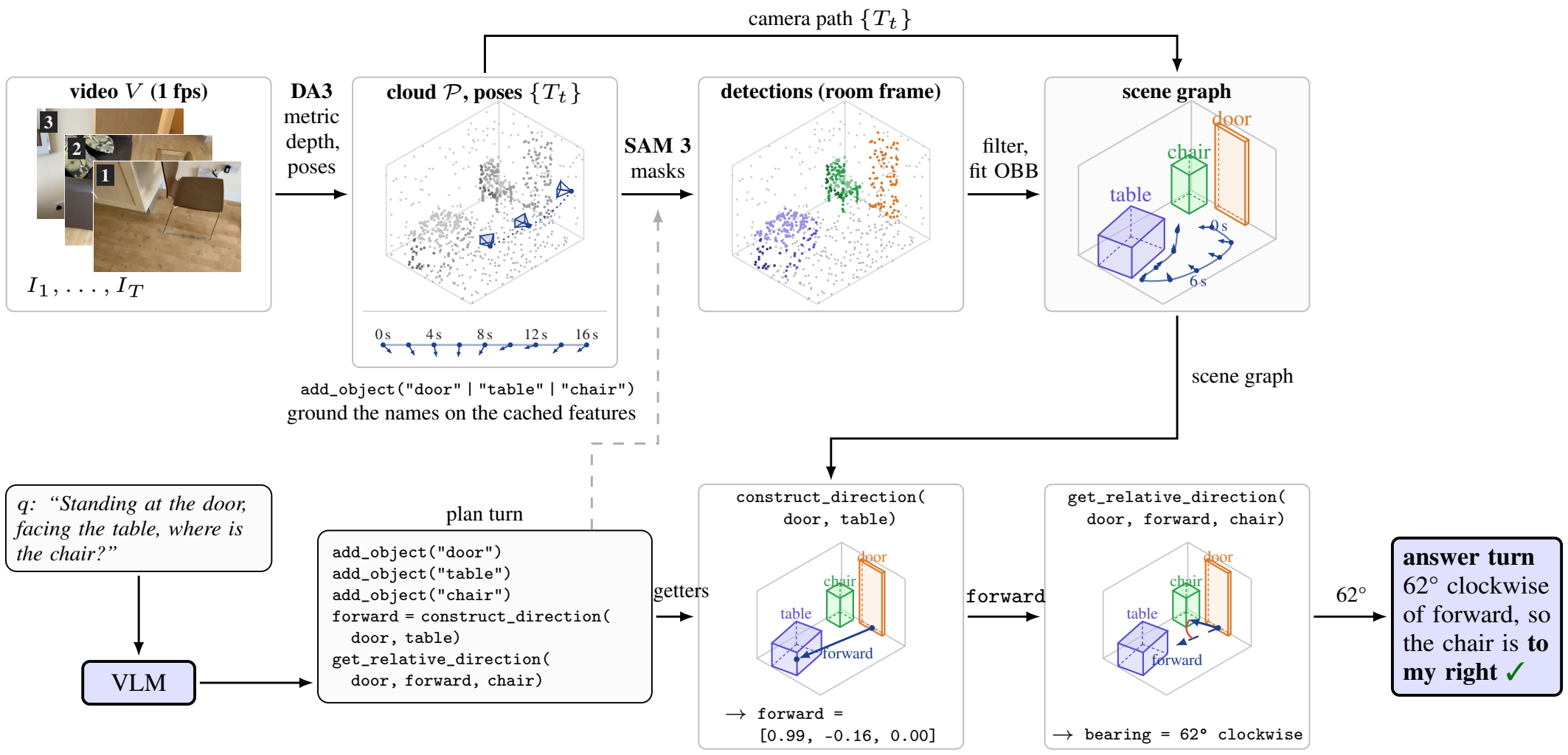

We are given an indoor video $V = (I_1, \ldots, I_T)$ and a
natural-language question $q$. SpaceMCP builds the scene graph
\emph{statically} from $V$ alone (Sec.~\ref{sec:method-static}),
expands it \emph{dynamically} once the VLM sees $q$
(Sec.~\ref{sec:method-dynamic}), and exposes both through a single MCP
interface (Sec.~\ref{sec:method-mcp}). Figure~\ref{fig:overview}
summarises the pipeline.

\subsection{Static generation}
\label{sec:method-static}

The static generation runs once per video and produces every artefact
that is independent of the question: a metric world-frame point cloud,
the floor plane and a 2D room frame. SAM~3's vision tower is also run
on every frame here and its activations cached, so per-question
\texttt{add\_object} calls (Sec.~\ref{sec:method-dynamic}) only invoke
the much cheaper text-conditioned grounding head.

We sample $V$ at 1~fps (reducing sampling for videos over 200~s) and
feed the frames to Depth Anything 3 \citep{da3}, a feed-forward 3D
transformer that returns \emph{metric} per-pixel depth, per-frame
intrinsics and extrinsics, and a confidence map in a single pass.
High-confidence pixels are unprojected into a fused world-frame point
cloud $\mathcal{P} \in \mathbb{R}^{N\times 3}$ in metres. To put
$\mathcal{P}$ in a 2D room frame, we run SAM~3 with the prompt
``floor'' over the cached features, project the masked pixels into
world coordinates, and fit a plane $\Pi$ with RANSAC \citep{ransac}.
With $\hat{n}$ the plane normal and $c$ its centroid, we build an
orthonormal basis $(\hat u, \hat v, \hat n)$ and rotate $\mathcal{P}$
so the floor lies in the $XY$ plane: $p_\text{room} = R_\text{room}(p
- c)$ with $R_\text{room} = [\hat u\;\hat v\;\hat n]^\top$. A 2~cm
occupancy grid over $XY$ (cells with at least 10 points are occupied)
is fit with a weighted-IoU rectangle, sweeping its rotation over $\pm
90^\circ$ around the AABB orientation, to give the four metric corners
$\mathcal{C} \in \mathbb{R}^{4\times 2}$ and room area
$A_\text{room}$. End-to-end, static generation takes $8.34$~s on an
Nvidia B200 for a 60-second clip, about $7.2\times$ faster than real
time.

\subsection{Dynamic querying}
\label{sec:method-dynamic}

Once $q$ is known, the VLM names the objects it needs and the scene
graph executes the corresponding \texttt{add\_object} /
\texttt{add\_objects} calls (Sec.~\ref{sec:method-mcp}). The two differ
only in their instance assumption. \texttt{add\_object(name)} assumes
\texttt{name} is a \emph{single} physical object: every pixel SAM~3
attributes to it in any frame is pooled and one oriented bounding box
(OBB) is fitted to the union. \texttt{add\_objects(name)} assumes
\emph{several} instances may exist: per-frame, per-instance masks are
kept separate, cross-frame correspondence is resolved by a greedy
mutex clustering (Appx.~\ref{app:impl}) under the constraint that two
detections from the same frame belong to different objects, and an
OBB is fitted per cluster. The API documentation directs the VLM to
\texttt{add\_object} for size, distance, direction and order questions
and to \texttt{add\_objects} for counting; instances enter the graph
under the noun phrase as label.

\paragraph{Grounding.}
Each call runs SAM~3's grounding head against the cached activations
with the noun phrase as prompt, using a confidence cascade
($0.85 \!\to\! 0.7 \!\to\! 0.5 \!\to\! 0.4 \!\to\! 0.3$) that falls to
lower thresholds when a pass returns no detections. Pixels above the
chosen threshold are unprojected via the cached DA3 depth and
transformed into the metric room frame. A raw detection is noisy ---
mask boundaries bleed onto neighbouring surfaces, where depth is least
reliable --- so before fitting an OBB we discard the least confident
15\% of points per detection by depth confidence and \verify{remove
residual outliers with DBSCAN \citep{dbscan}, keeping the dominant
cluster}. The OBB is fitted to the union of masks over all frames,
so an object never fully observed yields an under-estimated extent.
Bare nouns (``chair'') are grounded reliably but descriptive
constraints (``the chair by the door'') are not --- SAM~3 returns every
match or none --- so the VLM is instructed to use the bare noun phrase,
and descriptive disambiguation remains a limitation of the grounder.

\paragraph{Error propagation.}
If SAM~3 returns no instance at any cascade level, the object is absent
from the graph, every getter referencing it fails, and the system
falls back to the direct answer (Sec.~\ref{sec:method-mcp}); if it
returns a contaminated mask, the OBB is wrong but well-formed and the
error propagates silently into every downstream getter, with the
answer-turn plausibility check as the only safeguard. Because only the
grounding head runs here, a four-prompt augmentation completes in
$6.50$~s, and getter calls never touch the backbones again.

\begin{figure}[t]
\centering
\newcommand{\sgpanel}[1]{{\setlength{\fboxsep}{0pt}\setlength{\fboxrule}{0.4pt}%
  \color{black!35}\fbox{#1}}}
\setlength{\tabcolsep}{2pt}
\renewcommand{\arraystretch}{0.9}
\begin{tabular}{@{}ll@{\hspace{10pt}}ll@{}}
\sgpanel{\includegraphics[height=0.18\linewidth]{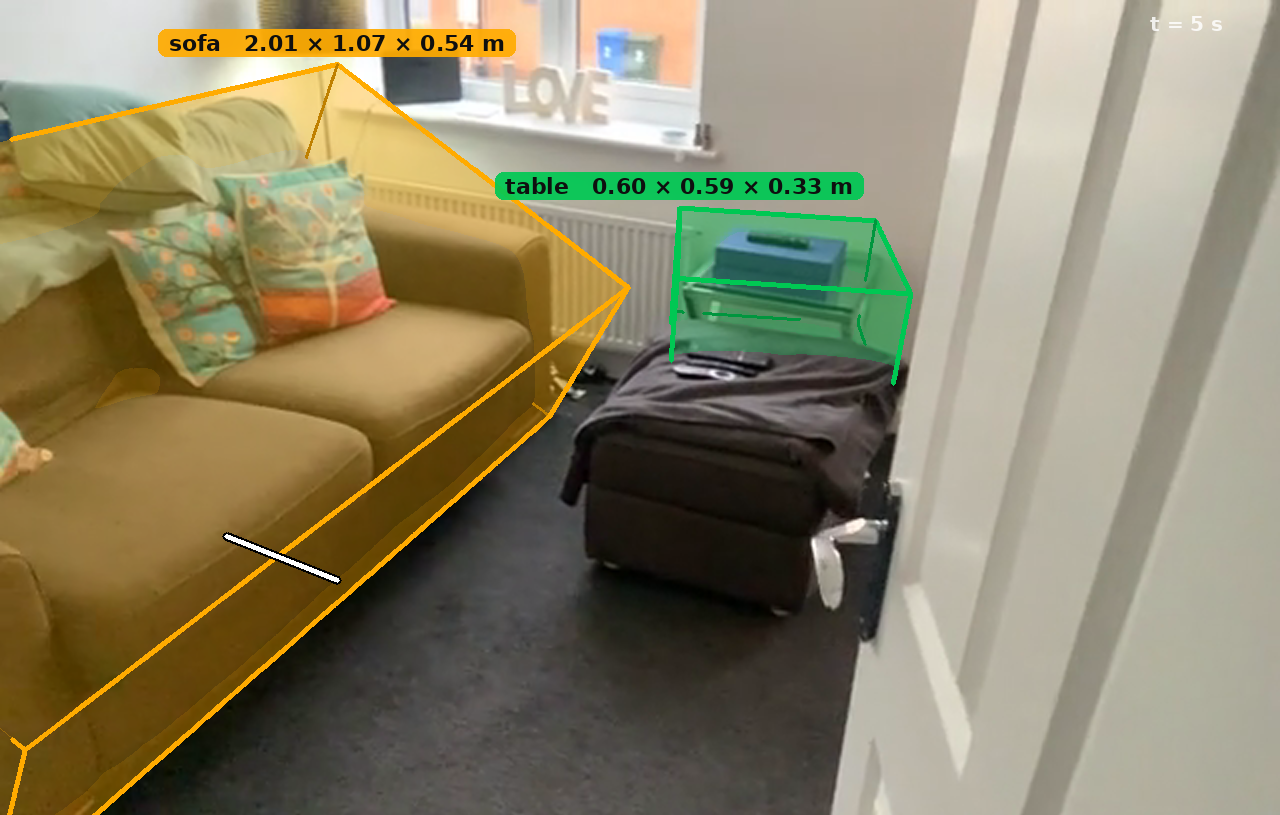}} & \sgpanel{\includegraphics[height=0.18\linewidth]{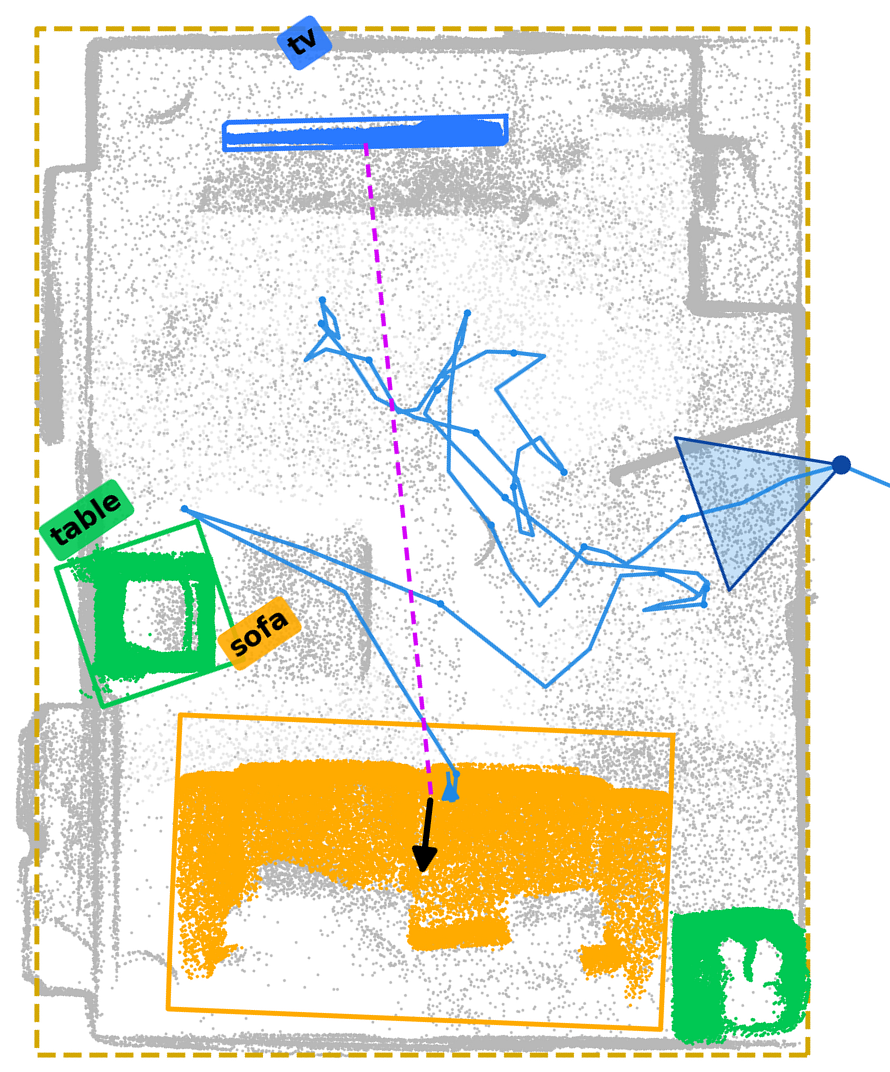}} &
\sgpanel{\includegraphics[height=0.18\linewidth]{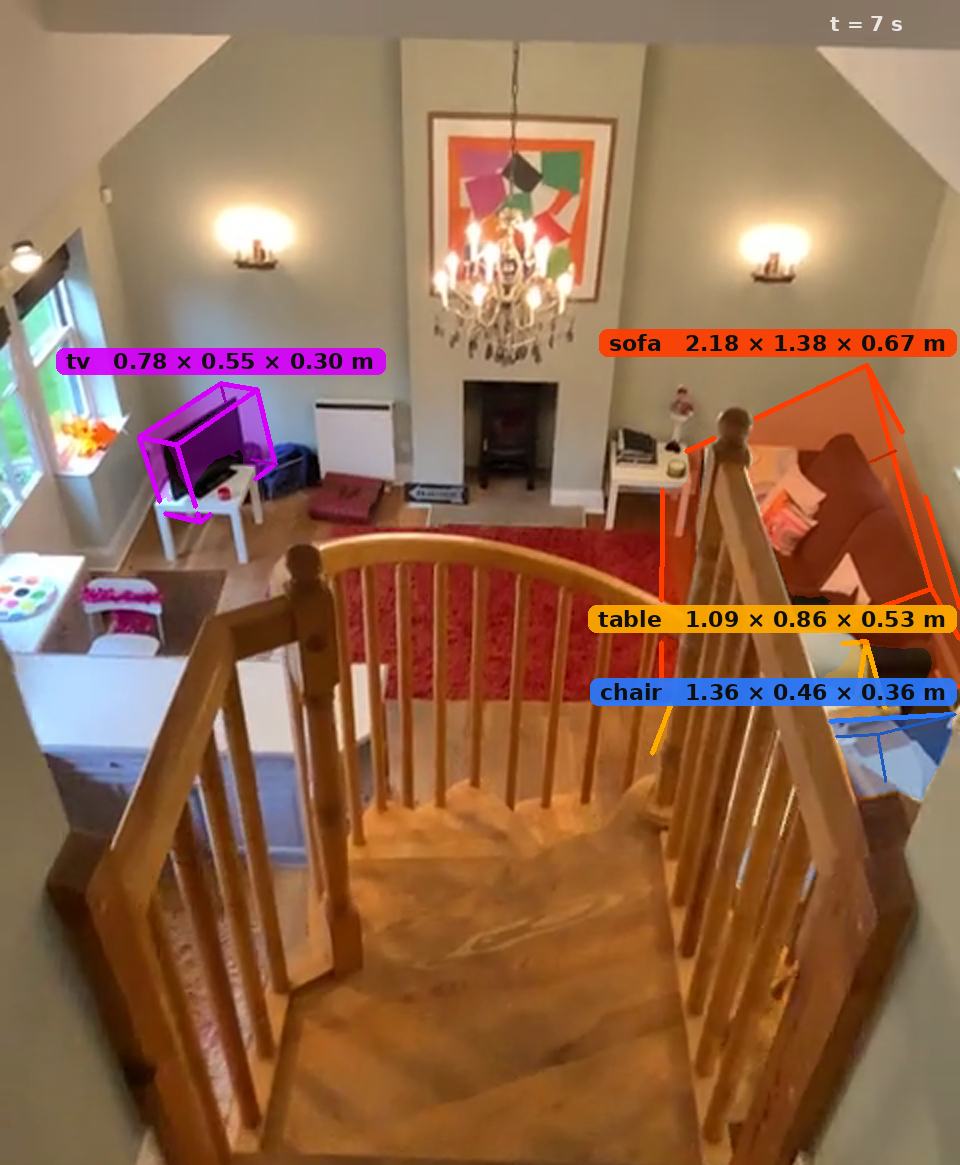}} & \sgpanel{\includegraphics[height=0.18\linewidth]{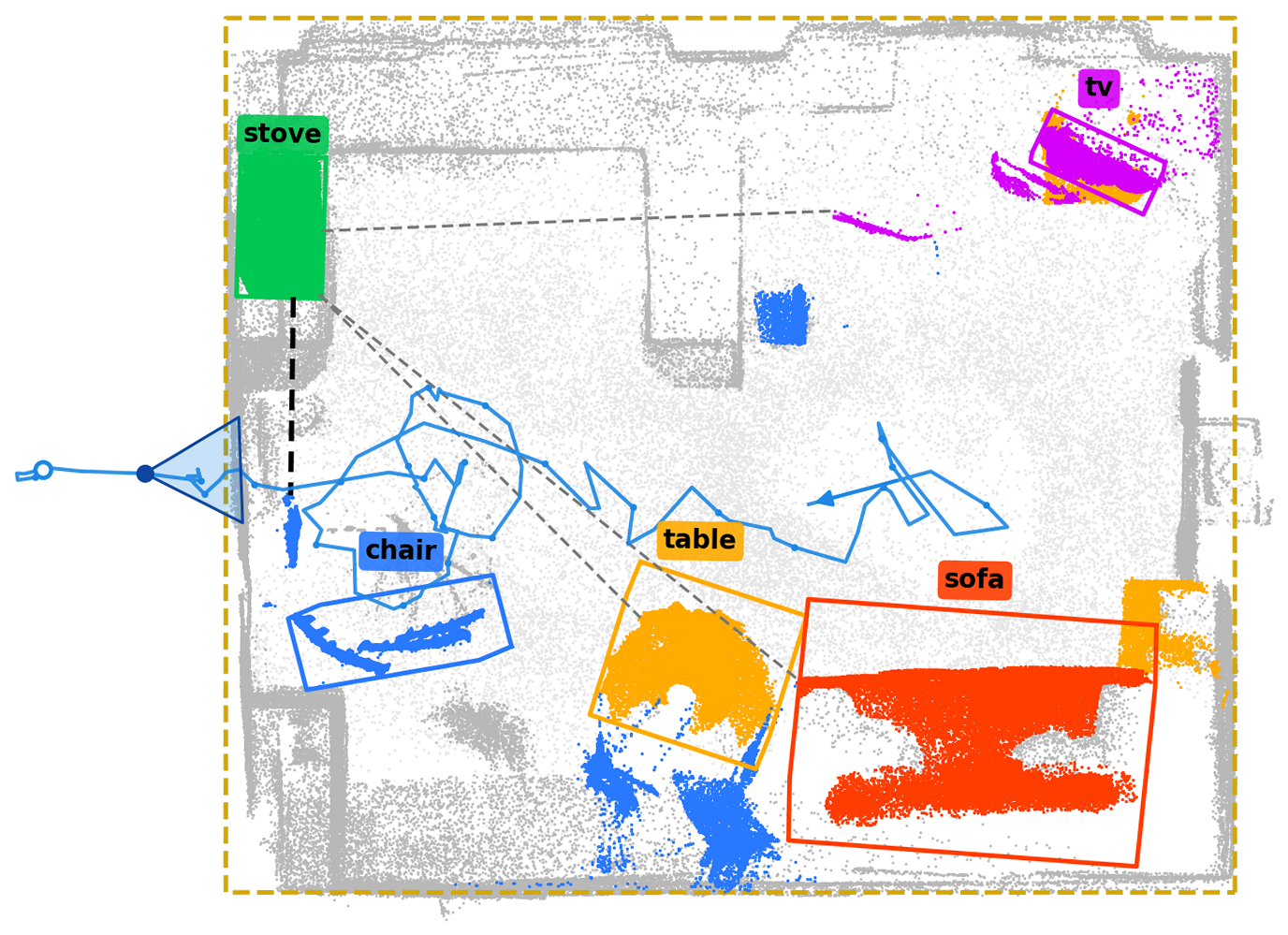}} \\
\end{tabular}
\caption{\textbf{Scene-graph outputs} for two videos. Each pair shows
a frame with the recovered OBBs and their metric extents, then the
room-frame map with the point cloud, room rectangle, object
footprints, camera path and the geometry returned for the question.}
\label{fig:scenegraph}
\end{figure}

\subsection{The MCP scene-graph interface}
\label{sec:method-mcp}

We expose the scene graph via a Model-Context-Protocol \citep{mcp}
server with two augmentation tools and a read-only query API in three
groups: \emph{static room geometry}, \emph{object geometry}, and
\emph{camera and pose operations} (Table~\ref{tab:api}, Appx.~\ref{app:api});
Fig.~\ref{fig:scenegraph} shows the resulting graph for two videos. Each instance
stores an OBB and its segmented point cloud: the OBB powers coarse
queries (size, dimensions, volume), the point cloud precise ones such
as \texttt{get\_distance}, which returns the closest-pair distance
between two clouds. Variables are supported --- a camera position from
\texttt{get\_camera\_pose}, or a direction from
\texttt{construct\_direction}, can be fed into
\texttt{get\_relative\_direction} --- so a single plan composes
arbitrary call sequences without an extra round trip.

Given $q$, the VLM is shown the current scene-graph state (room
footprint and labels already in the graph) with the API
documentation, and emits a single JSON plan combining the augmentation
calls naming the targets it needs with the getter calls whose values
answer $q$. The plan is executed and the model is shown the returned
values in a second answer turn that asks for the answer in the question's expected unit; that
answer is the final prediction. The
documentation instructs the agent to use the exact noun phrase from
the question and \emph{not} to use the interface for questions that
hinge on common sense rather than geometry (e.g.\ route planning),
which are answered directly. If the plan contains no augmentation
calls, an LLM-driven \emph{target extractor} pass --- one short prompt
that only identifies the objects named in $q$ --- supplies them.

\paragraph{Direct-answer fallback.}
Alongside the plan, the VLM also produces a direct answer to $q$ from
$V$. When the plan fails outright --- the model fails to produce a
valid API call (a malformed plan or an unknown tool), or SAM~3 returns
no instance for a requested object at any cascade level --- we report
the direct answer instead. The fallback fires on 2.5\% of VSI-Bench,
14.3\% of MMSI-Bench and 5.0\% of VSTI-Bench questions
(Sec.~\ref{sec:attribution}).

\section{Experiments}
\label{sec:experiments}

\subsection{Setup}
\label{sec:setup}

\paragraph{Benchmarks and metrics.}
We evaluate on VSI-Bench \citep{vsibench} (counting, size, room size,
distance, direction, route plan, appearance order), MMSI-Bench
\citep{mmsi} (multi-image spatial reasoning over disjoint viewpoints)
and VSTI-Bench \citep{vlm3r} (spatio-temporal queries over evolving
viewpoints: camera direction and displacement, camera--object and
object--object relations). Multiple-choice questions are scored by
exact letter match; numeric questions by Mean Relative Accuracy,
$\mathrm{MRA} = \frac{1}{10}\sum_{\theta \in \{0.5, 0.55, \ldots,
0.95\}} \mathbf{1}\!\left(|\hat{y} - y|/y < 1 - \theta\right)$, which
credits a prediction $\hat y$ whose relative error to the ground truth
$y$ falls below each of ten thresholds.

\paragraph{Baselines.}
We compare against Gemini 3.1 Flash Lite Preview~\citep{gemini31}, the base model of
our SpaceMCP, and recent open-source models post-trained for spatial
reasoning: SpaceR~\citep{spacer}, Cambrian-S-7B~\citep{cambrian},
VST-7B (RL)~\citep{vst}, SpatialReasoner~\citep{spatialreasoner}, and
SenseNova-SI-1.1 (InternVL3-8B)~\citep{sensenova}, the current SOTA on
VSI and MMSI, plus VLAA-Thinker-3B/7B~\citep{vlaathinker}, which is
post-trained for general rather than spatial reasoning. We
run all released checkpoints ourselves on the same test set. Until
now, training-free approaches \citep{mindjourney,tiger,mssr} have
mostly been outperformed by post-training methods, with some of the
best results by any approach coming from models such as
SenseNova-SI \citep{sensenova} and VST-7B; we therefore compare against
these post-trained models rather than prior training-free
tool-augmented methods. Because Gemini's size and training
data are undisclosed, we also report SpaceMCP on two open-source
backbones of known size, Qwen3.5-9B and Qwen3.6-27B \citep{qwen35,qwen36},
each with and without the interface (Sec.~\ref{sec:attribution}).

\paragraph{Runtime cost.}
Static generation takes $8.34$~s per 60-second clip on one B200,
amortised over all questions on that video; a four-prompt augmentation
takes $6.50$~s, getters run in milliseconds, and the VLM sees two
turns instead of one --- far below
methods whose inner loop is a generative novel-view or world-model
rollout \citep{mindjourney,spatialdreamer}.

\subsection{Main results}
\label{sec:results}

Tables~\ref{tab:vsi}--\ref{tab:mmsi} report per-task accuracies and
means; the \textbf{SpaceMCP (ours)} row is driven by
Gemini~3.1~Flash~Lite~Preview, and Sec.~\ref{sec:attribution} reports
open-source Qwen backbones. SpaceMCP obtains the best mean on all
three benchmarks: 71.8 on VSI-Bench ($+8.9$ over the strongest
baseline, SenseNova-SI), 49.2 on MMSI-Bench ($+12.8$) and 71.1 on
VSTI-Bench ($+19.2$). The largest gains are on sub-tasks that reduce
to a metric quantity the interface returns directly: absolute
distance (88.3 vs.\ at most 53.4 for any baseline), relative direction
(82.8 vs.\ 69.0), camera displacement (92.1 vs.\ 40.3) and
camera--object absolute distance (67.9 vs.\ 62.8). Baselines retain
the lead on a few sub-tasks, most visibly VSI-Bench counting (46.2
vs.\ 82.1 for VST-7B), analysed in Sec.~\ref{sec:attribution}.

\begin{table}[t]
\caption{VSI-Bench accuracy (\%). AppOrd: appearance order;
RelDir$^{*}$: relative direction (three difficulty sub-tasks, weighted
3$\times$ in the Mean); RelDist / AbsDist: relative / absolute
distance; Route: route plan; Size: object size; Room: room size.}
\label{tab:vsi}
\centering
\scriptsize
\setlength{\tabcolsep}{4pt}
\begin{tabular*}{\linewidth}{@{\extracolsep{\fill}}l@{\hspace{\tabcolsep}\hspace{\fill}\vrule\hspace{\tabcolsep}}cccccccc@{\hspace{\tabcolsep}\hspace{\fill}\vrule\hspace{\tabcolsep}}c@{}}
\toprule
Model & AppOrd & RelDir$^{*}$ & RelDist & Route & AbsDist & Count & Size & Room & Mean \\
\midrule
SpaceR               & 55.2 & 44.8 & 20.7 & 37.9 & 43.8 & 77.2 & 79.3 & 52.4 & 50.1 \\
Cambrian-S-7B        & 58.6 & 51.7 & 51.7 & 44.8 & 41.0 & 61.4 & \textbf{86.9} & 55.5 & 55.5 \\
VST-7B               & 62.1 & 51.7 & 44.8 & 44.8 & 34.1 & \textbf{82.1} & 82.4 & 52.1 & 55.8 \\
SenseNova-SI         & 65.5 & 69.0 & 51.7 & \textbf{55.2} & 53.4 & 51.0 & 79.3 & \textbf{65.5} & 62.9 \\
SpatialReasoner      & 24.1 & 37.9 & 24.1 & 37.9 & 43.1 & 37.9 & 52.1 &  5.9 & 33.9 \\
VLAA-Thinker-7B      & 37.9 & 32.2 & 20.7 & 44.8 & 12.1 & 34.5 & 61.4 & 44.1 & 35.2 \\
VLAA-Thinker-3B      & 17.2 & 47.1 & 27.6 & 51.7 & 30.3 & 24.1 & 21.4 & 26.6 & 34.0 \\
\midrule
\textbf{SpaceMCP (ours)} & \textbf{69.0} & \textbf{82.8} & \textbf{69.0} & 51.7 & \textbf{88.3} & 46.2 & 82.8 & 62.8 & \textbf{71.8} \\
\bottomrule
\end{tabular*}
\end{table}

\begin{table}[t]
\caption{VSTI-Bench accuracy (\%). CamDir: camera direction; CD:
camera displacement; COAD: camera--object absolute distance;
CO-D1/D2/D3: camera--object distance at three difficulty levels;
OO-Pos: object--object position. Mean$_5$ excludes CD and COAD (the
protocol of Sec.~\ref{sec:attribution}).}
\label{tab:vsti}
\centering
\scriptsize
\setlength{\tabcolsep}{4pt}
\begin{tabular*}{\linewidth}{@{\extracolsep{\fill}}l@{\hspace{\tabcolsep}\hspace{\fill}\vrule\hspace{\tabcolsep}}ccccccc@{\hspace{\tabcolsep}\hspace{\fill}\vrule\hspace{\tabcolsep}}cc@{}}
\toprule
Model & CamDir & CD & COAD & CO-D1 & CO-D2 & CO-D3 & OO-Pos & Mean & Mean$_5$ \\
\midrule
SpaceR               & 34.5 & 29.0 & 41.7 & 27.6 & 51.7 & 72.4 & 72.4 & 47.0 & 51.7 \\
Cambrian-S-7B        & 48.3 & 39.3 & 35.9 & 58.6 & \textbf{69.0} & 69.0 & 37.9 & 51.1 & 56.6 \\
VST-7B               & 24.1 &  9.7 & 47.9 & 58.6 & 58.6 & 75.9 & 62.1 & 48.1 & 55.9 \\
SenseNova-SI         & 44.8 & 40.0 & 47.6 & \textbf{69.0} & 58.6 & \textbf{79.3} & 24.1 & 51.9 & 55.2 \\
SpatialReasoner      & 34.5 & 40.3 & 62.8 & 24.1 & 51.7 & 75.9 & 72.4 & 51.7 & 51.7 \\
VLAA-Thinker-7B      & 17.2 &  2.8 & 14.8 & 31.0 & 58.6 & 69.0 & 72.4 & 38.0 & 49.7 \\
VLAA-Thinker-3B      & 44.8 & 23.8 & 16.2 & 20.7 & 51.7 & 65.5 & 41.4 & 37.7 & 44.8 \\
\midrule
\textbf{SpaceMCP (ours)} & \textbf{58.6} & \textbf{92.1} & \textbf{67.9} & 62.1 & 65.5 & 65.5 & \textbf{86.2} & \textbf{71.1} & \textbf{67.6} \\
\bottomrule
\end{tabular*}
\end{table}

\begin{table}[t]
\caption{MMSI-Bench accuracy (\%). Mot: motion; PR: pairwise
relations; Attr: attributes; MSR: multi-step reasoning.}
\label{tab:mmsi}
\centering
\scriptsize
\setlength{\tabcolsep}{3pt}
\resizebox{\linewidth}{!}{%
\begin{tabular}{lccccccccccc|c}
\toprule
Model & Mot-C & Mot-O & PR-CC & PR-CO & PR-CR & PR-OO & PR-OR & PR-RR & Attr-A & Attr-M & MSR & Mean \\
\midrule
SpaceR                 & 17.2 & 34.5 & 34.5 & 41.4 & 34.5 & 13.8 & 37.9 & 13.8 & 20.7 & 17.2 & 17.2 & 25.7 \\
Cambrian-S-7B          & 10.3 & 27.6 & 37.9 & 51.7 & 34.5 & 24.1 & 31.0 & 27.6 & 13.8 & 20.7 & \textbf{31.0} & 28.2 \\
VST-7B                 & 27.6 & 41.4 & 37.9 & 34.5 & 27.6 & 37.9 & \textbf{44.8} & 20.7 & 37.9 & 17.2 & 27.6 & 32.3 \\
SenseNova-SI           & 27.6 & 44.8 & \textbf{44.8} & 48.3 & 48.3 & \textbf{51.7} & 37.9 & \textbf{41.4} & 10.3 & 27.6 & 17.2 & 36.4 \\
SpatialReasoner        & 20.7 & 41.4 & 34.5 & 13.8 & 24.1 & 13.8 & 20.7 & 27.6 & 31.0 & 24.1 & 24.1 & 25.1 \\
VLAA-Thinker-7B        & 13.8 & 37.9 & 41.4 & 24.1 & 34.5 & 17.2 & 41.4 & 17.2 & 27.6 & 27.6 & 13.8 & 27.0 \\
VLAA-Thinker-3B        & 13.8 & 24.1 & 24.1 & 27.6 & 31.0 & 27.6 & 34.5 & 31.0 & 20.7 & 20.7 & 24.1 & 25.4 \\
\midrule
\textbf{SpaceMCP (ours)}         & \textbf{55.2} & \textbf{48.3} & \textbf{44.8} & \textbf{69.0} & \textbf{65.5} & 41.4 & 31.0 & \textbf{41.4} & \textbf{51.7} & \textbf{65.5} & 27.6 & \textbf{49.2} \\
\bottomrule
\end{tabular}}
\end{table}

\subsection{Ablation study}
\label{sec:attribution}

We ablate how the geometry is presented to the VLM (callable interface
versus the same facts as flattened text) and the VLM backbone, and
analyse the counting failure. Prompts and
questions are held fixed; VSTI numbers here are Mean$_5$ of
Table~\ref{tab:vsti}.

\paragraph{Interface versus flattened geometry.}
To separate the contribution of the interface from that of the
perception models, we hold the perception outputs fixed and vary only
how they reach the VLM. \emph{Flattened (positions)} instantiates the
objects named in the question through the same SAM~3 / DA3 path and
serialises their 3D centres as text in the prompt. \emph{Flattened
(scene facts)} pre-evaluates every getter whose arguments are fully
determined by the named objects and the room (room geometry, every
per-object dimension, count and first-appearance frame, and every
pairwise distance; Appx.~\ref{app:impl}). \verify{Objects are named by the
target-extractor prompt of Sec.~\ref{sec:method-mcp}}, so in both
variants the VLM issues no calls: it receives the facts alongside the
video and answers in one turn, as SpatialPIN-style text injection does
\citep{spatialpin}. Only the compositional getters (relative direction
from a constructed forward vector; camera-pose operations on a frame
index) cannot be pre-evaluated, since their argument space cannot be
enumerated compactly.

Table~\ref{tab:interface} gives the result. The full serialisation
recovers most of the gain on VSTI-Bench (65.5 vs.\ 67.6) but leaves
10.1 points on VSI-Bench and 7.8 on MMSI-Bench, and the position-only
serialisation is worse than no geometry at all on MMSI-Bench (14.8
vs.\ 34.3). Part of the gain is thus simply access to metric
perception, which a good static presentation captures; the remainder
is the interface: the VLM asks for what it needs rather than searching
a wall of numbers, composes primitives (a forward direction, then a
bearing) that no static dump contains, and is not exposed to
irrelevant values. The MMSI collapse under the naive serialisation
shows that \emph{how} geometry is presented is not a detail.

\begin{table}[h]
\caption{Ablations (accuracy, \%). \emph{Top:} presentation of the
same SAM~3 / DA3 outputs to the Gemini backbone. \emph{Bottom:} each
backbone with and without SpaceMCP. $\Delta$ relative to the direct
VLM; VSTI$_5$ is the five-category mean (all seven categories for the
Gemini pair: 49.0 vs.\ 71.1).}
\label{tab:interface}
\centering
\footnotesize
\setlength{\tabcolsep}{5pt}
\begin{tabular*}{\linewidth}{@{\extracolsep{\fill}}lcccccc@{}}
\toprule
Presentation of perception outputs & VSI & $\Delta$ & MMSI & $\Delta$ & VSTI$_5$ & $\Delta$ \\
\midrule
None (direct VLM)                       & 55.8 & ---     & 34.3 & ---     & 52.3 & --- \\
Flattened text: object positions only  & 61.9 & $+6.1$  & 14.8 & $-19.5$ & 53.8 & $+1.5$ \\
Flattened text: full scene facts       & 61.7 & $+5.9$  & 41.4 & $+7.1$  & 65.5 & $+13.2$ \\
SpaceMCP (callable interface)          & \textbf{71.8} & $+16.0$ & \textbf{49.2} & $+14.9$ & \textbf{67.6} & $+15.3$ \\
\midrule
Qwen3.5-9B: direct / + SpaceMCP  & 54.8 / \textbf{72.6} & $+17.8$ & 35.0 / \textbf{39.9} & $+4.9$ & \textbf{56.9} / 53.8 & $-3.1$ \\
Qwen3.6-27B: direct / + SpaceMCP & 61.6 / \textbf{71.8} & $+10.2$ & 32.9 / \textbf{43.4} & $+10.5$ & 58.5 / \textbf{76.9} & $+18.4$ \\
\bottomrule
\end{tabular*}
\end{table}

\paragraph{VLM backbone.}
\label{sec:exp-ablation}
Table~\ref{tab:interface} (bottom) pairs each backbone with its
SpaceMCP-augmented counterpart under a fixed protocol. Gains on
VSI-Bench are large for all three ($+16.0$, $+17.8$, $+10.2$); on
MMSI-Bench they are largest for Gemini ($+14.9$); on VSTI-Bench
Qwen3.6-27B gains the most of any configuration ($+18.4$) while
Qwen3.5-9B loses 3.1 points. The fully open-source Qwen3.6-27B +
SpaceMCP reaches 71.8 / 43.4 / 76.9 and exceeds every open-source
baseline in Tables~\ref{tab:vsi}--\ref{tab:mmsi}, so the headline
result does not depend on the proprietary backbone. The API is
structured function calls and JSON, so driving it is easy for general
LLMs, with one caveat: the backbone must be a competent tool user.
Qwen3.5-9B's regression matches a failure mode we observed directly on
an earlier dataset, where it fell into circular re-planning and blew
its answer-turn budget, which we did not see on Qwen3.6-27B or Gemini
(Sec.~\ref{sec:limitations}).

\paragraph{Counting.}
\label{sec:counting}
Counting is SpaceMCP's weakest task (46.2 on VSI-Bench Count vs.\ 82.1
for VST-7B). Unlike every other sub-task it depends on cross-frame
instance association (\texttt{add\_objects}): a fixed $0.5$~m merge
threshold over-splits large objects seen from disjoint viewpoints and
over-merges adjacent small ones, and a single grounding miss changes a
count directly rather than merely perturbing a point set. Label noise
contributes (Appx.~\ref{app:labels}) but does not explain the gap. An
extent-relative threshold, a SAM~2/3-style tracker \citep{sam2,sam3}
deciding association by temporal continuity, or routing counting to
the direct answer are natural remedies we leave to future work.

\subsection{Limitations}
\label{sec:limitations}

SpaceMCP inherits the failure modes of its backbones: DA3 degrades on
reflective, transparent or texture-poor surfaces and returns
confidently wrong depth that propagates silently, and SAM~3 struggles
with descriptive constraints. Counting suffers from fixed-threshold
clustering (Sec.~\ref{sec:counting}), and on 2.5\% / 14.3\% / 5.0\% of
VSI / MMSI / VSTI questions the plan fails and the direct answer is
returned. Reliable use needs a backbone that follows the plan/answer
protocol: a 27B open model sufficed, a 9B model regressed on one
benchmark, and plan validation, budget caps and loop detection are
not yet implemented. The room getters assume a visible floor and one
rectangular, static room; outdoors they should be disabled and metric
depth is less accurate at range, while the object and camera getters
carry no indoor assumption. We have not evaluated outdoor, open-space
or dynamic settings. \texttt{construct\_direction} needs two named
points, so questions about a single object's intrinsic orientation
lack an anchor (orientation estimation \citep{orientanything} could
plug in as a getter), and the pipeline has no plausibility prior, so
gross outliers are caught only by the VLM's answer turn. Labels we
could not reconcile with the scene are flagged, not excluded
(Appx.~\ref{app:labels}).

\section{Discussion and Conclusion}
\label{sec:conclusion}

The metric heavy lifting in SpaceMCP is done by perception, but the
VLM is more than a router: given identical perception outputs it does
markedly better when it can \emph{ask} than when it is \emph{told}
(Table~\ref{tab:interface}): it chooses which quantities matter,
composes primitives no static representation contains, and
sanity-checks the returned values. Because the interface decouples
\emph{what} the VLM reasons about from \emph{how} values are computed,
better reconstructors or grounders can be swapped in without touching
the VLM. The same capability turns indoor video into a metric scene
graph of objects and people, so deployment should be gated by consent
and retention limits.

We presented SpaceMCP, a training-free, tool-augmented framework that
endows a VLM with metric spatial reasoning by constructing a scene
graph in real time and exposing it through a generic MCP interface. It
delivers strong, format-uniform results on VSI-Bench, MMSI-Bench and
VSTI-Bench, outperforming the strongest proprietary VLMs and the
largest-scale spatially trained models \citep{spatialvlm,sensenova},
and ablations show the gains do not
depend on the proprietary backbone and are only partly recovered by
handing the VLM the same geometry as text. We frame SpaceMCP as an
initial proposal for a standard interface between VLMs and 3D
perception.

\bibliographystyle{plainnat}

\appendix

\section{The MCP Interface}
\label{app:api}

Table~\ref{tab:api} lists every tool exposed to the VLM.

\begin{table}[h]
\caption{The SpaceMCP MCP interface. Sizes in metres, areas in
m\textsuperscript{2}, volumes in m\textsuperscript{3}, angles in
degrees; bearings use the $[0,360)$ compass convention
(0\textdegree=ahead, 90\textdegree=right), pose-pair rotations and
pitch are signed. Position arguments accept an object name (its OBB
centre), a coordinate, or a \texttt{"\%var.path"} reference into an
earlier result; direction arguments accept a
\texttt{construct\_direction} result or a camera heading.}
\label{tab:api}
\centering
\tiny
\setlength{\tabcolsep}{3pt}
\begin{tabular*}{\linewidth}{@{\extracolsep{\fill}}ll@{\hspace{0.6em}}ll@{}}
\toprule
\textbf{Tool} & \textbf{Description} & \textbf{Tool} & \textbf{Description} \\
\midrule
\multicolumn{2}{l}{\emph{Augmentation}}                                  & \multicolumn{2}{l}{\emph{Object geometry (cont.)}} \\
\texttt{add\_object(name)}            & Segment one instance.            & \texttt{get\_object\_volume(n)}                  & Sum of OBB volumes. \\
\texttt{add\_objects(name)}           & All instances, clustered.        & \texttt{get\_distance($n_1, n_2$)}              & Closest-pair distance. \\
\multicolumn{2}{l}{\emph{Static room geometry}}                          & \texttt{get\_first\_appearance(n)}              & First visible frame. \\
\texttt{get\_room\_size()}            & Area, long, short side.          & \texttt{construct\_direction($p_1, p_2$)}       & Unit direction $p_1\!\to\!p_2$. \\
\texttt{get\_room\_corners()}         & Four $[x,y]$ corners.            & \texttt{get\_relative\_direction($p_1, \vec{f}, p_2$)} & Bearing of $p_2$ from $p_1$. \\
\texttt{get\_room\_center()}          & 3D room centre.                  & \multicolumn{2}{l}{\emph{Camera and pose operations}} \\
\texttt{get\_room\_axes()}            & Long/short/up axes.              & \texttt{get\_camera\_pose(frame)}               & Camera extrinsics. \\
\multicolumn{2}{l}{\emph{Object geometry}}                               & \texttt{rotate\_clockwise(pose, deg)}           & Pose rotated CW. \\
\texttt{get\_all\_object\_labels()}   & Labels in graph.                 & \texttt{rotate\_anticlockwise(pose, deg)}       & Pose rotated CCW. \\
\texttt{get\_object\_count(n)}        & Number of instances.             & \texttt{move\_camera\_forward(pose, m)}         & Pose translated. \\
\texttt{get\_object\_shortest\_dim(n)}& Smallest OBB extent.             & \texttt{get\_distance\_to\_object(pose, n)}     & 2D distance. \\
\texttt{get\_object\_middle\_dim(n)}  & Middle OBB extent.               & \texttt{get\_translation\_between($p_a, p_b$)} & Forward/right delta. \\
\texttt{get\_object\_longest\_dim(n)} & Largest OBB extent.              & \texttt{get\_rotation\_between($p_a, p_b$)}    & Signed yaw delta. \\
\texttt{get\_object\_height(n)}       & Vertical extent.                 & \texttt{get\_pitch\_between($p_a, p_b$)}       & Signed pitch delta. \\
\texttt{get\_object\_height\_above\_ground(n)} & Height above floor.     & \texttt{get\_camera\_displacement($p_a, p_b$)} & 3D L2 between positions. \\
\texttt{get\_object\_horizontal\_extents(n)}   & Footprint dims.          & & \\
\texttt{get\_object\_center(n)}       & 3D centroid.                     & & \\
\bottomrule
\end{tabular*}
\end{table}

\section{MCP Tool-Use Prompts}
\label{app:prompts}

The system prompts for stage (A) (direct video
QA) and the two turns of the scene-graph flow (plan and answer) each enumerate the available getters
(Table~\ref{tab:api}), demand a strict JSON output schema for the
plan turn, and include a small number of in-context examples drawn
from a held-out subset of VSI-Bench training videos.

\section{Additional Implementation Notes}
\label{app:impl}

\paragraph{Cross-frame instance clustering for \texttt{add\_objects}.}
SAM~3 emits per-frame, per-instance masks; we recover physical
objects by clustering those detections across frames. Each detection
is a 3-D point set in the room frame, obtained by back-projecting
the masked pixels via cached DA3 depth and filtering the bottom
$15\%$ by depth confidence (and dropping SAM~3 instances with score
$< 0.5$). For all cross-frame pairs $(i, j)$ we compute the
min-point-distance (subsampling to $300$ points per detection for
speed). We walk the pairs in ascending min-distance order and
greedily union pair $(i, j)$ whenever \emph{(i)}
$\text{min\_dist}(i, j) \le 0.5$~m and \emph{(ii)} the resulting
union would not place two same-frame detections in the same
component. The same-frame mutex is enforced \emph{transitively} via
component frame-sets, so a chain of below-threshold cross-frame
links cannot collapse two adjacent objects (e.g.\ neighbouring
chairs at a table) into one cluster. An OBB is then fit to each
resulting component. Sec.~\ref{sec:attribution} analyses the failure
modes of this procedure.

\paragraph{Fallback triggers.}
The direct answer replaces the plan's answer when (a) the plan turn
does not parse as the required JSON schema or names a tool that does
not exist, or (b) an augmentation call returns no instance at any
level of the confidence cascade, in which case every getter that
references the missing object fails. Getters that execute but return
implausible values do \emph{not} trigger the fallback; catching those
is left to the answer turn
(Sec.~\ref{sec:method-mcp}).

\paragraph{Flattened-text baselines.}
Both flattened variants of Sec.~\ref{sec:attribution} instantiate the
objects named in the question through the unchanged
\texttt{add\_object} / \texttt{add\_objects} path, so SAM~3 and DA3 see
exactly what they see under SpaceMCP. \emph{Positions only} serialises
one line per object, \texttt{<name>: center\_m=[x, y, z]}. \emph{Scene
facts} serialises, in order: the room block (area, long and short
side, corners, centre, axes); one block per object with its three OBB
extents, height, height above ground, horizontal extents, volume,
centre, instance count and first-appearance frame; and one line per
object pair with the closest-point distance. The text is placed in the
user turn before the question, the VLM sees the same sampled frames as
under SpaceMCP, and it answers in one turn with the direct-QA prompt.
No getter that takes a direction or a camera pose as an argument is
serialised, for the reason given in Sec.~\ref{sec:attribution}.

\paragraph{Caching.}
Per-video DA3 outputs and floor / wall SAM~3 masks are cached on
disk and re-used across questions sharing the same video, so the
amortised per-question cost is dominated by the per-target SAM~3
forwards triggered by the \texttt{add\_object} / \texttt{add\_objects}
calls in the plan turn.

\section{Benchmark Labels We Could Not Reconcile with the Scene}
\label{app:labels}

During analysis we encountered labels that we could not reconcile
with the visible scene. We list them here so that others can verify
them; none is excluded from any number reported in this paper, and we
note that comparisons against our own estimates cannot by themselves
establish that a label is wrong.

\textbf{VSTI-Bench, camera--object absolute distance.} Sample 1288
(label $2.1$~m; SpaceMCP measures $1.61$~m), sample 1929 (label
$0.8$~m; $0.38$~m) and sample 1512 (label $1.2$~m; $1.6$~m).

\textbf{VSI-Bench, counting.} Sample 156 is labelled with two chairs
while three are visible (SpaceMCP predicts 3); sample 22 is labelled
with two couches while one is visible (SpaceMCP predicts 1). In both
cases one of the post-trained baselines (SenseNova-SI) reproduces the
annotated count. We flag this as consistent with benchmark-format
overfitting but have not quantified it, and it does not account for
SpaceMCP's counting deficit (Sec.~\ref{sec:attribution}).

\end{document}

%% file: fig1_reasoning.tex
\resizebox{\linewidth}{!}{%
\begin{tikzpicture}[
    font=\footnotesize,
    >={Latex[length=2mm]},
    box/.style={draw, thick, rounded corners=2pt, align=center, inner sep=4pt},
    vlm/.style={box, fill=blue!12, minimum width=1.3cm, minimum height=0.7cm},
    bubble/.style={draw, rounded corners=4pt, align=left, inner sep=6pt,
                   text width=4.4cm, fill=gray!3, font=\scriptsize},
    bubblegood/.style={bubble, draw=green!55!black},
    bubblebad/.style={bubble, draw=red!65!black},
    mcp/.style={draw, thick, fill=orange!22, rounded corners=2pt, inner sep=5pt,
                font=\scriptsize\bfseries, minimum height=0.7cm, minimum width=0.9cm},
    arr/.style={->, thick, shorten >=1pt, shorten <=1pt}
]

\node[draw=none, inner sep=0pt] (vid2) at (0, 3.0) {%
\begin{tikzpicture}[x=1cm, y=1cm, baseline=(current bounding box.center)]
  \fill[black] (0,0) rectangle (5.4,1.7);
  \foreach \x in {0.05,0.40,0.75,1.10,1.45,1.80,2.15,2.50,2.85,3.20,3.55,3.90,4.25,4.60,4.95,5.30}{
    \fill[white] (\x,1.50) rectangle ({\x+0.13},1.62);
    \fill[white] (\x,0.08) rectangle ({\x+0.13},0.20);
  }
  \node[inner sep=0pt, anchor=south west] at (0.15,0.30)
        {\includegraphics[width=1.6cm, height=1.1cm]{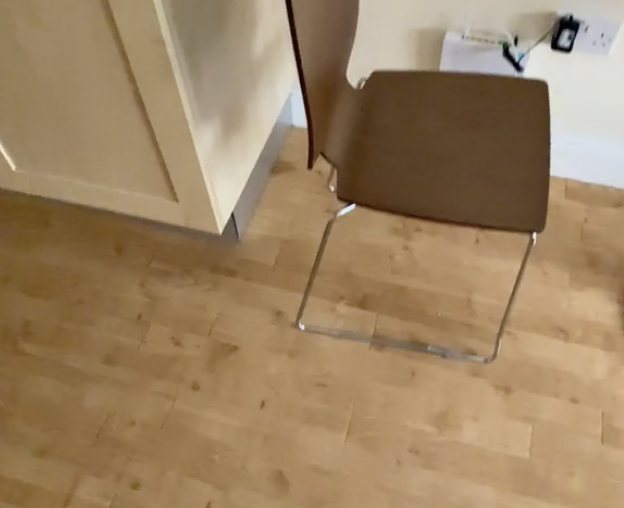}};
  \node[inner sep=0pt, anchor=south west] at (1.90,0.30)
        {\includegraphics[width=1.6cm, height=1.1cm]{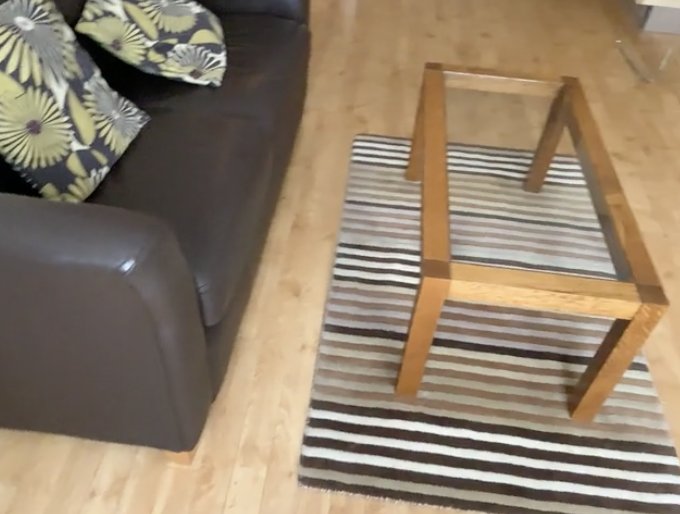}};
  \node[inner sep=0pt, anchor=south west] at (3.65,0.30)
        {\includegraphics[width=1.6cm, height=1.1cm]{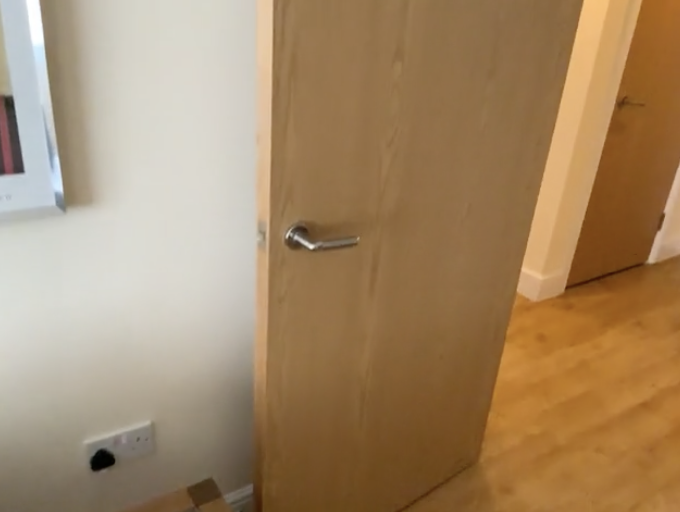}};
\end{tikzpicture}};
\node[below=2pt of vid2, align=center, font=\small\itshape, text width=4cm]
    (cap2) {``If I am standing at the door and facing the table, where is the chair relative to me?''};

\node[vlm] (vlmS2) at (7.5, 1.5) {VLM};
\node[above=0.6mm of vlmS2, font=\scriptsize\itshape] (lblS) {\textbf{SpaceMCP (ours)}};
\node[vlm] (vlmD2) at (12.5, 1.5) {VLM};
\node[above=0.6mm of vlmD2, font=\scriptsize\itshape] (lblD) {Direct};

\draw[arr] (vid2.east) -| ($(lblS.north)+(0,0.05)$);
\draw[arr] (vid2.east) -| ($(lblD.north)+(0,0.05)$);

\node[bubblegood, anchor=north] (tS) at (7.5, 0.5) {%
\textit{``This asks for the chair's bearing from the door's viewpoint
  while facing the table. Once I have all three objects in the scene
  graph I can query the relative direction~--- first I need to ground
  the forward direction\ldots''}\\[3pt]
\textcolor{gray!55}{\hrulefill}\\[1pt]
{\ttfamily\scriptsize
get\_object(["door", "table", "chair"]);\\
forward = construct\_direction(door, table);\\
get\_relative\_direction(door, forward, chair)}\\[3pt]
\textcolor{gray!55}{\hrulefill}\\[1pt]
\null\hfill\textbf{62\textdegree}\hfill\null\\[3pt]
\textcolor{gray!55}{\hrulefill}\\[1pt]
\textit{``The angle is 62\textdegree\ clockwise from forward,
  therefore the chair is \textbf{to my right}
  \textcolor{green!45!black}{\ding{51}}''}};
\draw[arr] (vlmS2.south) -- (tS.north);

\node[bubblebad, anchor=north] (tD) at (12.5, 0.5) {%
\textit{``From the frames the door sits on one back wall and the chair
  is across the room on the opposite wall, with the table out in
  the middle. Standing at the door and facing the table I'd be
  looking diagonally across the room, with the chair off to one
  side of my view. Based on the layout, the chair is on my left
  $\Rightarrow$ \textbf{to my left}
  \textcolor{red!70!black}{\ding{55}}''}};
\draw[arr] (vlmD2.south) -- (tD.north);

\node[inner sep=0pt] (graph2) at (0, -3.8)
    {\includegraphics[width=5.0cm]{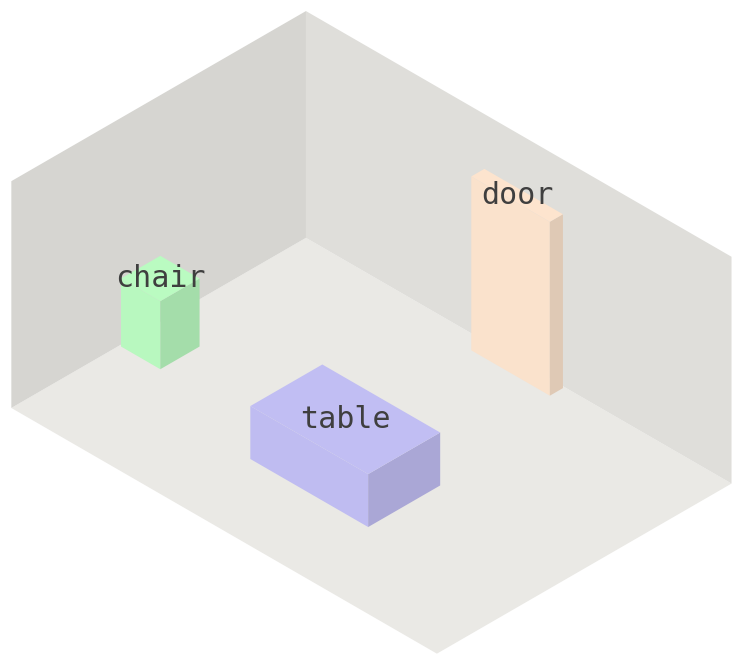}};

\draw[arr] (cap2.south) -- (graph2.north);

\node[mcp] (mcp2) at (4, -3.8) {MCP};

\draw[<->, thick, shorten >=1pt, shorten <=1pt] (graph2.east) -- (mcp2.west);

\coordinate (tool_start) at ($(tS.north west)+(0,-3.3)$);  
\coordinate (ang_start)  at ($(tS.north west)+(0,-5.3)$);  

\draw[arr] (tool_start) -| (mcp2.north);
\draw[arr] (mcp2.south) |- (ang_start);

\node[font=\scriptsize\itshape, above=0pt]
    at ($(tool_start)!0.5!(mcp2.north |- tool_start)$) {tool call};
\node[font=\scriptsize\itshape, below=0pt]
    at ($(ang_start)!0.5!(mcp2.south |- ang_start)$) {result};

\end{tikzpicture}%
}

%% file: fig2_overview.tex
\input{fig2_overview_macros.tex}%
\begin{tikzpicture}[font=\scriptsize, >={Latex[length=1.4mm]},
  isosmall/.style={x={(-0.2598cm,-0.15cm)}, y={(0.2598cm,-0.15cm)}, z={(0cm,0.3cm)}},
  call/.append style={minimum width=2.6cm, minimum height=2.6cm},
  ccode/.append style={font=\fontsize{5.2}{6.2}\selectfont\ttfamily}]

\node[panel, minimum width=2.6cm, minimum height=2.3cm] (p1) at (1.45,1.15) {};
\node[panel, minimum width=2.6cm, minimum height=2.85cm] (p2) at (4.85,0.875) {};
\node[panel, minimum width=2.6cm, minimum height=2.3cm] (p3) at (8.25,1.15) {};
\node[panel, minimum width=2.6cm, minimum height=2.3cm, fill=gray!3] (p4) at (11.65,1.15) {};
\node[ptitle] at (p1.north) {video $V$ (1 fps)};
\node[ptitle] at (p2.north) {cloud $\mathcal{P}$, poses $\{T_t\}$};
\node[ptitle] at (p3.north) {detections (room frame)};
\node[ptitle] at (p4.north) {scene graph};

\node[inner sep=0pt, draw=white, line width=0.6pt] at (1.17,1.45)
  {\includegraphics[width=1.44cm, height=1.08cm]{fig_frames/f3.png}};
\node[inner sep=0pt, draw=white, line width=0.6pt] at (1.45,1.19)
  {\includegraphics[width=1.44cm, height=1.08cm]{fig_frames/f2.png}};
\node[inner sep=0pt, draw=white, line width=0.6pt] at (1.73,0.93)
  {\includegraphics[width=1.44cm, height=1.08cm]{fig_frames/f1.png}};
\node[sm, anchor=south west, inner sep=1pt] at (0.3,0.1) {$I_1,\ldots,I_T$};
\node[chip, fill=gray!30!black] at (0.47,1.97) {3};
\node[chip, fill=gray!30!black] at (0.75,1.71) {2};
\node[chip, fill=gray!30!black] at (1.03,1.45) {1};

\begin{scope}[shift={(4.989cm,1.19cm)}, iso]
  \roomwire{gray!28, line width=0.35pt}
  \pgfmathsetseed{2026}
  \cloud{gray!60}{gray!45}{cGrayObj}{cGrayObj}{cGrayObj}{gray!40}
  \draw[cCam, densely dotted, line width=0.45pt]
    (0.3,2.35,1.3) .. controls (0.7,2.45,1.2) and (1.1,2.45,1.1) .. (1.5,2.35,1.05)
    .. controls (1.9,2.3,1.05) and (2.3,2.4,1.1) .. (2.6,2.35,1.1);
  \cam{(0.3,2.35,1.3)}{(0.5,-0.87,0)}{(-0.87,-0.5,0)}
  \cam{(1.5,2.35,1.05)}{(0,-1,0)}{(-1,0,0)}
  \cam{(2.6,2.35,1.1)}{(-0.5,-0.87,0)}{(-0.87,0.5,0)}
\end{scope}
\draw[gray!35, line width=0.35pt] (3.65,0.0) -- (6.05,0.0);
\begin{scope}[shift={(3.85cm,-0.33cm)}]
  \draw[cCam!45, line width=0.5pt] (0,0) -- (2.0,0);
  \foreach \px/\py/\dx/\dy [count=\i from 0] in \pathposes{%
    \draw[cCam, ->, >={Latex[length=0.6mm]}, line width=0.35pt] ({\i*0.25},0) -- ({\i*0.25+0.13*\dx},{0.13*\dy});
    \fill[cCam] ({\i*0.25},0) circle (0.7pt);
    \pgfmathtruncatemacro{\tsec}{2*\i}
    \ifodd\i\else
      \node[font=\fontsize{4.2}{5}\selectfont, gray!25!black, anchor=south, inner sep=0.6pt] at ({\i*0.25},0.03) {\tsec\,s};
    \fi}
\end{scope}

\begin{scope}[shift={(8.389cm,1.19cm)}, iso]
  \roomwire{gray!28, line width=0.35pt}
  \pgfmathsetseed{2026}
  \cloud{gray!55}{gray!42}{cTable}{cChair}{cDoor}{gray!38}
\end{scope}

\begin{scope}[shift={(11.789cm,1.19cm)}, iso]
  \roomwire{gray!45, line width=0.4pt}
  \isopath
  \node[font=\fontsize{4.5}{5}\selectfont, cCam, anchor=west, inner sep=0.8pt] at (0.7,1.15,0.06) {0\,s};
  \node[font=\fontsize{4.5}{5}\selectfont, cCam, anchor=north, inner sep=0.8pt] at (1.9,2.1,0) {6\,s};
  \obb{\tc}{0.6}{0.4}{0.75}{cTableL}{cTable}
  \obb{\cc}{0.25}{0.25}{0.95}{cChairL}{cChair}
  \obb{\dc}{0.05}{0.425}{2.05}{cDoorL}{cDoor}
  \node[sm, cTable, anchor=south, inner sep=1pt] at \tc{-0.6}{-0.4}{0.75} {table};
  \node[sm, cChair, anchor=south, inner sep=1pt] at \cc{0}{0}{1.0} {chair};
  \node[sm, cDoor, anchor=south, inner sep=1pt] at \dc{0}{0}{2.1} {door};
\end{scope}

\draw[arr] (p1.east) -- (p2.west |- p1.east);
\node[alab, text width=0.75cm] at (3.15,1.27) {\textbf{DA3}\\ metric depth, poses};
\draw[arr] (p2.east |- p3.west) -- (p3.west);
\node[alab, text width=0.75cm] at (6.55,1.27) {\textbf{SAM 3}\\ masks};
\draw[arr] (p3.east) -- (p4.west);
\node[alab, text width=0.75cm] at (9.95,1.27) {filter,\\ fit OBB};
\draw[arr] (p2.north) -- ++(0,0.4) -| (p4.north);
\node[sm, anchor=south, inner sep=1pt] at (8.25,2.72) {camera path $\{T_t\}$};

\node[bubble, anchor=north, text width=2.4cm] (qb) at (1.45,-1.7)
  {$q$: ``Standing at the door, facing the table, where is the chair?''};
\node[vlm] (vlmn) at (1.45,-3.65) {VLM};
\draw[arr] (qb.south) -- (vlmn.north);
\node[code, text width=3.0cm, inner sep=4pt, font=\fontsize{5}{6}\selectfont\ttfamily] (plan) at (4.85,-3.0)
  {add\_object("door")\\ add\_object("table")\\ add\_object("chair")\\
   forward = construct\_direction(\\ \ \ \ door, table)\\ get\_relative\_direction(\\ \ \ \ door, forward, chair)};
\node[sm, anchor=south, inner sep=1pt] at (plan.north) {plan turn};
\draw[arr] (vlmn.east) -- (plan.west |- vlmn.east);

\draw[darr] (5.9,-2.18) -- (5.9,-1.3) -| (6.55,1.05);
\node[sm, align=right, anchor=east, inner sep=1pt] at (6.38,-0.9)
  {{\fontsize{5}{6}\selectfont\ttfamily add\_object("door"\,|\,"table"\,|\,"chair")}\\ ground the names on the cached features};

\draw[arr] (p4.south) |- (8.25,-1.25) -- (8.25,-1.7);
\node[sm, anchor=west, inner sep=1pt] at (11.75,-0.65) {scene graph};

\node[call] (c1) at (8.25,-3.0) {};
\node[ccode, anchor=north, align=center] at (c1.north) {construct\_direction(\\ \quad door, table)};
\begin{scope}[shift={(8.354cm,-2.93cm)}, isosmall]
  \roomwire{gray!45, line width=0.35pt}
  \obb{\tc}{0.6}{0.4}{0.75}{cTableL}{cTable}
  \obb{\cc}{0.25}{0.25}{0.95}{cChairL}{cChair}
  \obb{\dc}{0.05}{0.425}{2.05}{cDoorL}{cDoor}
  \node[font=\fontsize{4.5}{5}\selectfont, cDoor, anchor=south, inner sep=0.5pt] at \dc{0}{0}{2.1} {door};
  \node[font=\fontsize{4.5}{5}\selectfont, cChair, anchor=south, inner sep=0.5pt] at \cc{0}{0}{1.0} {chair};
  \node[font=\fontsize{4.5}{5}\selectfont, cTable, anchor=south, inner sep=0.5pt] at \tc{-0.6}{-0.4}{0.75} {table};
  \fill[cCam] (0.05,1.2,0.02) circle (0.8pt);
  \fill[cCam] (2.5,0.8,0.02) circle (0.8pt);
  \draw[dir] (0.05,1.2,0.02) -- (2.5,0.8,0.02)
    node[pos=0.32, below=-1pt, font=\fontsize{4.5}{5}\selectfont, cCam] {forward};
\end{scope}
\node[ccode, anchor=south, align=left] at (c1.south) {$\to$ forward =\\ \phantom{$\to$} [0.99, -0.16, 0.00]};

\node[call] (c2) at (11.65,-3.0) {};
\node[ccode, anchor=north, align=center] at (c2.north) {get\_relative\_direction(\\ \quad door, forward, chair)};
\begin{scope}[shift={(11.754cm,-2.93cm)}, isosmall]
  \roomwire{gray!45, line width=0.35pt}
  \obb{\tc}{0.6}{0.4}{0.75}{cTableL}{cTable}
  \obb{\cc}{0.25}{0.25}{0.95}{cChairL}{cChair}
  \obb{\dc}{0.05}{0.425}{2.05}{cDoorL}{cDoor}
  \node[font=\fontsize{4.5}{5}\selectfont, cDoor, anchor=south, inner sep=0.5pt] at \dc{0}{0}{2.1} {door};
  \node[font=\fontsize{4.5}{5}\selectfont, cChair, anchor=south, inner sep=0.5pt] at \cc{0}{0}{1.0} {chair};
  \node[font=\fontsize{4.5}{5}\selectfont, cTable, anchor=south, inner sep=0.5pt] at \tc{-0.6}{-0.4}{0.75} {table};
  \fill[cCam] (0.05,1.2,0.02) circle (0.8pt);
  \draw[dir, dashed, line width=0.5pt] (0.05,1.2,0.02) -- (1.5,0.963,0.02)
    node[pos=0.92, below=-1pt, font=\fontsize{4.5}{5}\selectfont, cCam] {forward};
  \draw[dir] (0.05,1.2,0.02) -- (0.36,0.3,0.02);
  \draw[cDrop, line width=0.5pt] (0.8888,1.0626,0.02) arc (-9.3:-71:0.85);
\end{scope}
\node[ccode, anchor=south] at (c2.south) {$\to$ bearing = 62\textdegree\ clockwise};

\node[vlm, anchor=west, text width=1.45cm, align=left, minimum height=1.2cm] (ans) at (13.75,-3.0)
  {\textbf{answer turn}\\ 62\textdegree\ clockwise of forward, so the chair is \textbf{to my right} \textcolor{green!45!black}{\ding{51}}};

\draw[arr] (plan.east |- c1.west) -- (c1.west);
\node[alab, text width=0.45cm] at (6.73,-2.9) {getters};
\draw[arr] (c1.east) -- (c2.west);
\node[alab, text width=0.75cm] at (9.95,-2.9) {\texttt{forward}};
\draw[arr] (c2.east) -- (ans.west);
\node[alab, text width=0.5cm] at (13.35,-2.9) {62\textdegree};

\end{tikzpicture}%